\documentclass{article}

\usepackage[preprint]{neurips_2026}

\author{
Kejia Chen$^{1}$,
Jiawen Zhang$^{1}$,
Yihong Wu$^{2}$,
Kewei Gao$^{1}$,\\
\textbf{Jian Lou$^{3}$,
Zunlei Feng$^{1}$,
Mingli Song$^{1}$,
Ruoxi Jia$^{4}$}
\\
$^{1}$Zhejiang University \quad
$^{2}$Université de Montréal\quad
$^{3}$Sun Yat-sen University \quad
$^{4}$Virginia Tech
}

\usepackage[utf8]{inputenc} 
\usepackage[T1]{fontenc}    
\usepackage[table]{xcolor}
\usepackage{amsmath,amsthm,amssymb,booktabs,comment,graphicx,subfig,multirow,bbm,tikz,enumitem,algorithm,algorithmic,stackengine,makecell,nicefrac,array,setspace,tabularx,latexsym,times,varwidth,arydshln,threeparttablex,pgf,wrapfig,lipsum,colortbl,float,microtype,pifont,siunitx,dsfont,hyperref}
\usepackage[inkscapelatex=false]{svg}
\usepackage[many]{tcolorbox}
\usepackage[textsize=tiny]{todonotes}
\usepackage{caption}
\usepackage{wrapfig}
\usepackage{wrapstuff}
\hypersetup{
 colorlinks=true,
 urlcolor=magenta,
 citecolor=gray!66!darkgreen,
}

\newcommand{\projectname}{\texttt{CASPO}}

\definecolor{darkgreen}{rgb}{0.0, 0.5, 0.0} 
\definecolor{darkred}{rgb}{0.5, 0.0, 0.0}

\theoremstyle{plain}

\theoremstyle{definition}

\theoremstyle{remark}

\title{Confidence-Aware Alignment Makes Reasoning LLMs More Reliable}

\begin{document}

\maketitle

\begin{abstract}
  Large reasoning models often reach correct answers through flawed intermediate steps, creating a gap between final accuracy and reasoning reliability. Existing alignment strategies address this with external verifiers or massive sampling, limiting scalability. In this work, we introduce \projectname~(\underline{C}onfidence-\underline{A}ware \underline{S}tep-wise \underline{P}reference \underline{O}ptimization), a framework that aligns token-level confidence with step-wise logical correctness through iterative Direct Preference Optimization, without training a separate reward model. During inference, we propose Confidence-aware Thought (CaT), which leverages this calibrated confidence to dynamically prune uncertain reasoning branches with negligible $O(V)$ latency. Experiments across ten benchmarks and multiple model families show that \projectname~ consistently improves reasoning reliability and inference efficiency. Notably, \projectname~ scales to Qwen3-8B-Base and surpasses tree-search baselines on AIME'24 and AIME'25 without using reward-model data. We also release a step-wise dataset with confidence annotations to support fine-grained analysis of reasoning reliability. Code is available at ~\url{https://github.com/Thecommonirin/CASPO}.
\end{abstract}

\section{Introduction}
\label{sec.introduction}

Large reasoning models (LRMs) such as OpenAI-o1 \citep{jaech2024openai} and Qwen-3 \citep{yang2025qwen3} have substantially advanced mathematical and scientific problem-solving through detailed step-by-step generation. However, optimizing these models purely for final-answer correctness masks a critical vulnerability: they frequently arrive at correct conclusions via logically flawed intermediate steps \citep{arcuschin2503chain}. In high-stakes domains such as medicine and finance \citep{fadeeva2024fact, zhang2025towards}, relying on invalid reasoning traces poses significant risks. Therefore, reliable LRM deployment demands not only accurate final outputs but verifiably sound reasoning trajectories.

The root cause of this vulnerability lies in a fundamental misalignment between a model's internal confidence and logical correctness. In current LRMs, token-level probabilities reflect superficial string fluency and pattern frequency rather than true deductive validity \citep{arcuschin2503chain, yang2025probability}. Consequently, a model might confidently hallucinate a syntactically valid but logically incorrect step, while exhibiting low confidence when executing a rigorous but unfamiliar derivation. This pervasive miscalibration prevents internal confidence from serving as a reliable metric for self-verification.

Current efforts to improve reliability mainly operate at the trajectory level. Chain-of-Thought (CoT) \citep{wei2022chain} elicits intermediate steps through prompting, Self-Consistency \citep{wang2022self} aggregates multiple paths via majority voting, and reinforcement learning frameworks such as Group Relative Policy Optimization (GRPO) align models with preferred trajectories using verifiable rewards \citep{guo2025deepseek}. Even scaling methods such as rStar-Math \citep{guan2025rstar} largely treat the reasoning process as a monolithic output. This trajectory-centric paradigm presents a dilemma: trajectory-level methods overlook the reliability of individual steps, while search-intensive approaches incur computational costs that limit scalability.

To address this granularity gap, recent work introduces step-wise supervision to improve intermediate reasoning quality. Step-wise preference optimization \citep{razghandi2025cer} and process-based self-rewarding frameworks \citep{tu2025enhancing} integrate intermediate feedback into training, and weakness-driven augmentation strategies such as SwS \citep{liang2025sws} diagnose systematic failures. However, these methods typically rely on heuristic feedback or external verifiers and do not explicitly model the model's own uncertainty. Parallel efforts on confidence estimation via token probabilities \citep{xu2024genarm} face a further obstacle: empirical evidence \citep{arcuschin2503chain,yang2025probability,hu2025open} indicates that token-level confidence reflects surface fluency or frequent patterns rather than reasoning reliability. Models often assign high probability to syntactically correct but logically irrelevant steps, and underestimate uncertainty in complex derivations. Closing this gap requires a principled way to synchronize internal confidence with reasoning correctness.

Our core insight is that reliable reasoning requires calibration, where high predictive confidence is reserved for valid logical steps. Aligning internal probability with external correctness allows the model's own entropy to serve as a high-fidelity, zero-cost signal for guiding generation, removing the dependency on external evaluators during inference. Building on this principle, we propose \projectname~ (\underline{C}onfidence-\underline{A}ware \underline{S}tep-wise \underline{P}reference \underline{O}ptimization), a unified framework that operationalizes step-wise confidence across both training and inference.

During training, \projectname~ calibrates the model by constructing preference pairs that contrast correct but uncertain steps with confidently wrong predictions. These pairs are optimized via iterative DPO, aligning the model's probability distribution with logical validity. During inference, we introduce the Confidence-aware Thought (CaT) strategy, which uses cumulative step-wise confidence to dynamically expand promising paths and prune uncertain trajectories. This two-stage design propagates step-wise improvements into faithful final answers with negligible computational overhead.

In summary, our contributions are as follows: We propose \projectname~, a unified framework that uses intrinsic model confidence to achieve reliable reasoning without external verifiers. By aligning token-level entropy with logical correctness during training, the method enables self-calibration and addresses the tension between exploration and reliability. This calibration supports our CaT strategy, which prunes uncertain reasoning branches at inference with $O(V)$ latency overhead. Extensive experiments across ten benchmarks show consistent improvements with strong data and compute efficiency: \projectname~ raises the average accuracy of Qwen2.5-7B-Instruct from 44.4\% to 50.6\% and reaches 56.1\% with CaT at inference. On Qwen3-8B-Base, it surpasses tree-search baselines such as rStar-Math \citep{guan2025rstar} and Satori \citep{shen2025satori} on AIME2024 and AIME2025 without using any reward model data.
\section{Related Work}
\label{sec.related}

\textbf{Large Reasoning Models.} 
The evolution of LRMs has progressed from simple prompting to more sophisticated strategies. CoT showed that explicit step-by-step reasoning improves performance on complex tasks, while Self-Consistency \citep{wang2022self} enhanced robustness by aggregating multiple reasoning paths. Recent systems such as OpenAI’s o1 \citep{jaech2024openai} and DeepSeek-R1 \citep{guo2025deepseek} now leverage post-training to elicit extended reasoning traces for superior transparency and accuracy. In parallel, distillation techniques \citep{hsieh2023distilling} transfer high-quality reasoning trajectories to smaller models for efficiency. For instance, \citep{guan2025rstar} explicitly utilizes rationales from large teacher models to supervise smaller students, reducing data requirements while maintaining performance. Structured approaches such as Tree-of-Thoughts \citep{yao2023tree}, Graph-of-Thoughts \citep{besta2024graph}, and reinforcement learning \citep{zhang2024rest,zhang2025process,li2025treepo} further expand the reasoning space, albeit often at the expense of considerable computational efficiency.

\textbf{Reasoning Process Verification.}
As reasoning traces lengthen, ensuring their faithfulness becomes paramount. One prominent direction involves Process Reward Models (PRMs) \citep{lightman2023let,wang2023math}, trained on datasets such as PRM800K \citep{lightman2023let}, to score intermediate reasoning steps. Subsequent works such as PURE \citep{cheng2025stop} refine step-wise credit assignment in reinforcement learning. Beyond direct scoring, collaborative deliberation \citep{patnaik2025helps,patnaik2025learning} and selective rationale optimization \citep{lightman2023let,du2023improving,qu2024recursive} demonstrate that models can enhance reliability through mutual verification and preference ranking. While autonomous self-correction remains difficult, combining self-verification with lightweight external supervision offers a promising path toward reliability without the prohibitive cost of massive reward models.

\textbf{Verification-Enhanced Reasoning.}
Beyond evaluation, recent work integrates verification directly into reasoning. Test-time scaling generates multiple candidate solutions and selects the most reliable one, improving accuracy at high computational cost. At training time, reinforcement learning with verifiable rewards (e.g., SimpleRL \citep{zeng2025simplerl}, PURE \citep{cheng2025stop}) iteratively refine reasoning by rewarding faithful traces. To reduce reliance on explicit reward models, DPO-based methods approximate reward signals via likelihood estimation. While co-training generators and verifiers \citep{ouyang2022training} has also been explored, scalability and stability issues persist. Grounded in these directions, \projectname~ differs from these collaborative or external-distillation approaches: rather than relying on multi-model collaboration or mimicking teacher preferences, it unifies training and inference through \textit{intrinsic step-wise confidence calibration}, utilizing the student's own token entropy to guide reliable reasoning paths.




\section{Method}
\label{sec.analysis}

\begin{figure*}[!t]
    \centering
    \small
    \includegraphics[width=\linewidth]{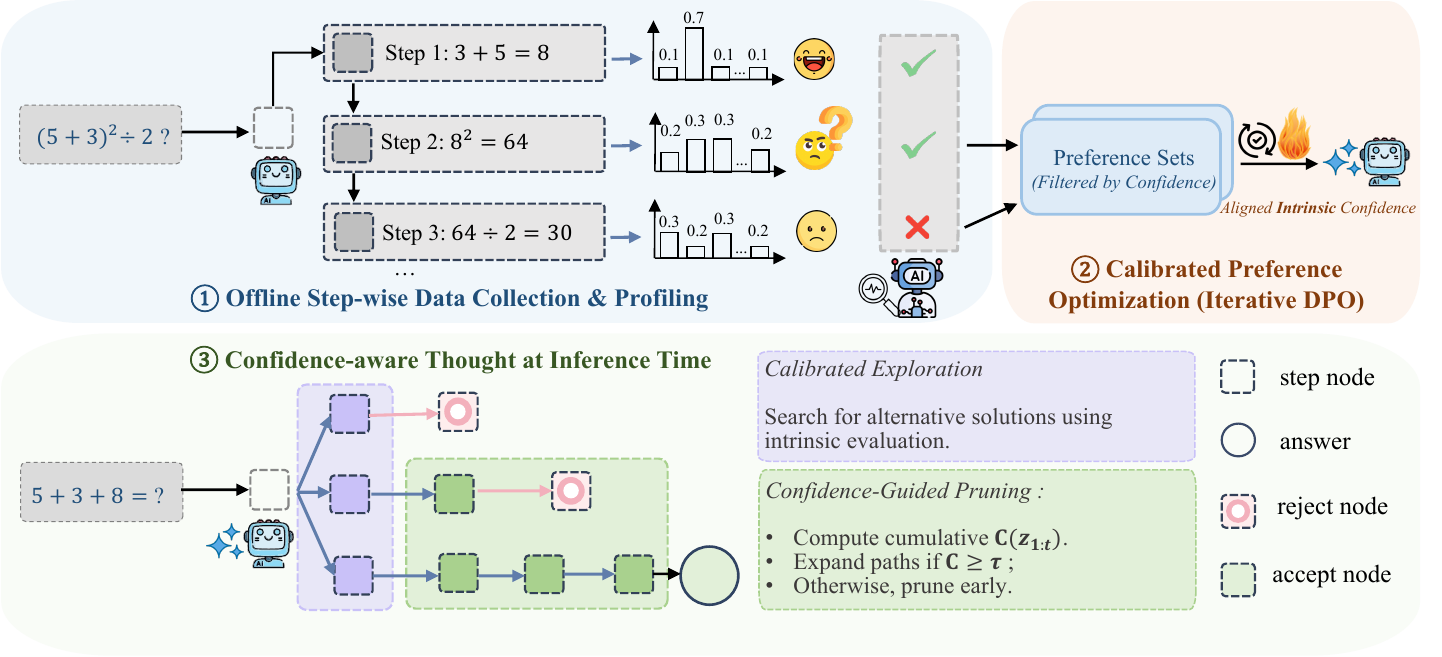}
    \caption{Overview of \projectname: A Unified Framework for Calibrated Reasoning. \projectname~first aligns intrinsic uncertainty with step-wise correctness through iterative preference optimization, then utilizes this calibrated confidence to dynamically prune reasoning trees during inference.}
    \vspace{-10pt}
    \label{fig:caspo}
\end{figure*}

\projectname~integrates intrinsic confidence estimation into a unified pipeline for both training and inference. As illustrated in Figure~\ref{fig:caspo}, our framework operates in two interconnected phases: (i) Confidence-Aware Preference Optimization, which aligns model uncertainty with step-wise correctness through iterative DPO, and (ii) Confidence-aware Thought (CaT) Inference, which leverages this calibrated uncertainty to dynamically navigate and prune the reasoning tree.

\subsection{Motivation and Problem Formulation}

Recent progress \citep{li2025treepo,wang2022self,zuo2025ttrl} in LRMs have highlighted a critical tension: sampling multiple reasoning paths boosts performance via diversity, but often introduces plausible yet hallucinated steps. Existing paradigms primarily rely on compute-intensive external verifiers or large-scale sampling, which introduce substantial inference overhead and provide limited insight into the model's intrinsic assessment of its own reasoning process.

Our goal is to equip the model with the ability to \textit{self-evaluate} the quality of each reasoning step $s_t$ conditioned on the current context $q_t$. We posit that genuine reasoning competence requires more than eventually arriving at the correct answer; it should also be reflected in the model's confidence when taking valid reasoning steps. In other words, correct reasoning should correspond to concentrated probability mass, or equivalently, low predictive entropy. \projectname~ therefore explicitly aligns the model's predicted probability distribution with the correctness of its reasoning steps, encouraging valid steps to be generated with high confidence while suppressing invalid or unreliable ones.

\subsection{~\projectname{}: \underline{C}onfidence-\underline{A}ware \underline{S}tep-wise \underline{P}reference \underline{O}ptimization}

\textbf{Notations.} We consider an auto-regressive language model $\pi_\theta$, which defines a next-token distribution $\pi_{\theta}(\cdot|x)$ given an input prompt $x$. For each query $x$ in the dataset $\mathcal{D}_\text{math}$, we view the reasoning process as a sequence of $m$ steps $s_{1:m} = (s_1, s_2, \ldots, s_m)$, leading to a final answer $a$. Each step $s_j$ is generated conditioned on a specific context, which we define as the \textbf{sub-question} $q_j$. This context concatenates the original query and the preceding reasoning history: 
\begin{equation}
    q_j = [x, s_1, s_2, \ldots, s_{j-1}],
\end{equation}
The model then generates the current step $s_j \sim \pi_{\theta} (\cdot |q_j)$. This formulation allows us to evaluate the quality of intermediate reasoning in a fine-grained manner.

\textbf{Step-wise Confidence Estimation.} To quantify the model's intrinsic uncertainty without external supervision, we utilize token-level entropy. Let the step answer $s_j$ generated by the model consist of a sequence of tokens $\{t_l\}_{l=1}^L$. The confidence of this specific step $s_j$ given context $q_j$ is computed as the negative average entropy:
\begin{equation} 
\small
\label{eq:confidence}
    \text{confidence}(s_j | q_j)
    =-\frac{1}{L}\sum_{l=1}^{L}\sum_{v\in \mathcal{V}}\pi_\theta(v|q_j, t_{<l}) \log \pi_\theta(v|q_j, t_{<l}),
\end{equation}
Where $L$ is the length of the step answer, $\mathcal{V}$ is the vocabulary, and $\pi_\theta(v|q_j)$ denotes the predictive distribution over tokens $v$. Higher cumulative entropy indicates greater uncertainty and, hence, lower confidence in the generation. We adopt token-level entropy as our uncertainty metric because it captures the model’s intrinsic uncertainty during generation, avoiding the overconfidence bias and hallucination sensitivity inherent in frequency-based diversity measures. This reference-free criterion evaluates each candidate’s confidence independently of the ground truth.

\textbf{Confidence-Aware Step-wise Data Collection.}
To obtain reliable supervision, we employ a strong model (e.g., Qwen2.5-Math-7B-Instruct) as an offline external evaluator. The evaluator verifies whether the step-wise answer $s_j$ is correct, and the model $\theta$ gives the confidence to the corresponding question $q_j$:
\begin{itemize}
    \item If $s_j$ is correct and has high confidence, omit it.
    \item If $s_j$ is correct but has low confidence, set $y_w=s_j$ and choose $y_l$ as a high-probability competing candidate step from $\pi_\theta(\cdot|q_j)$.
    \item If $s_j$ is incorrect, set $y_w$ to the correct answer and $y_l$ to $s_j$.
\end{itemize}
This selection strategy ensures that the preference dataset $\mathcal{D}$ consists exclusively of signals that drive the model towards calibrated correctness.

\textbf{Training: Confidence-Aware Preference Optimization.}
Based on the step-centric dataset $\mathcal{D}$ constructed in Algorithm~\ref{alg:caspo_compact}, we form preference pairs $(q_j, y_j^w, y_j^l)$. This design ensures that both reliable but uncertain predictions and confidently wrong predictions contribute to preference learning.

The training objective follows the DPO formulation, which encourages the target model $\pi_{\theta}$ to increase the relative likelihood of the preferred answer compared to the dispreferred one:
\begin{align}
    \mathcal{L}_{\text{DPO}} 
    = -\log \sigma \!\Bigg(
        \beta \left[ 
            \log \frac{\pi_{\theta}(y_j^w|q_j)}{\pi_{\text{ref}}(y_j^w|q_j)}
            - \log \frac{\pi_{\theta}(y_j^l|q_j)}{\pi_{\text{ref}}(y_j^l|q_j)}
        \right]
    \Bigg),
    \label{eq_preference}
\end{align}
where $\beta$ controls the strength of preference alignment.
To achieve continuous improvement, we adopt an Iterative DPO scheme: at each iteration $k$, the target model $\pi_{\theta_k}$ is optimized using the above loss with respect to the previous model $\pi_{\text{ref}}=\pi_{\theta_{k-1}}$ as the reference. After optimization, we set $\pi_{\text{ref}} \gets \pi_{\theta_k}$ for the next step. This allows the model to bootstrap its own reasoning capabilities, progressively refining both its accuracy and its confidence calibration.

\textbf{Inference: Confidence-aware Thought (CaT).} 
\label{sec:catot}
After iterative preference optimization, the model not only learns to prefer correct reasoning steps but also calibrates its confidence estimation at each step. This enables a CaT inference strategy: instead of committing to a single linear chain, the model explores a reasoning tree where each node corresponds to a partial reasoning trajectory $z_{1:t} = (z_1,\dots,z_t)$ with an associated confidence score
\begin{align}
    C(z_{1:t}) = \prod_{i=1}^t \text{confidence}(z_i|z_{1:i-1}),
\end{align}
where $\text{conf}(z_i|z_{1:i-1})$ denotes the normalized confidence of reasoning step $z_i$ given the previous context.  
During inference, a path is expanded only if its cumulative confidence $C(z_{1:t})$ exceeds a threshold $\tau$.
Low-confidence branches are pruned early, reallocating computational budget to more promising reasoning paths. This mechanism acts as an intrinsic \emph{self-correction} filter, ensuring that the final output is the result of a chain of high-confidence, valid reasoning steps.

\section{Experiments}
\label{sec.experiment}
We evaluate \projectname~ across multiple dimensions to verify its effectiveness in aligning reasoning confidence with correctness. Our analysis encompasses training comparisons, inference strategy scaling, out-of-domain generalization, and calibration quality. Comprehensive details
are in Appx.~\ref{apx:experi}.

\subsection{Settings}
\textbf{Models.} We employ Llama-3.1-8B-Instruct \citep{grattafiori2024llama}, Qwen2.5-Math-7B, and Qwen2.5-7B-Instruct \citep{yang2024qwen25} as our primary base models.
To verify scalability to stronger base models, we additionally conduct experiments on Qwen3-8B-Base \citep{yang2025qwen3}. For answer calibration during training data construction, Qwen2.5-Math-7B-Instruct serves as the evaluator.

\textbf{Baselines.}
We compare \projectname~ against two categories of methods: \textbf{(i) Training-based methods (Table~\ref{tbl:merged_performance})}, which update model parameters using verifiable self-improvement signals. We select six representative methods: GRPO \citep{shao2024deepseekmath}, Simple-RL-Zero \citep{zeng2025simplerl}, PURE-VR \citep{cheng2025stop}, rStar-Math \citep{guan2025rstar}, PCPO \citep{yang2025probability}, and DPO-VP \citep{tu2025enhancing}. For scalability comparisons, we also include tree-search-based methods rStar-Math \citep{guan2025rstar} and Satori \citep{shen2025satori}. \textbf{(ii) Inference-time methods (Table~\ref{tbl:inferencemath})}, which modify the decoding process without parameter updates. We compare against CoT \citep{kojima2022large}, Self-Consistency \citep{wang2022self}, and DiPT \citep{just2024dipt}. Detailed descriptions of these baselines are deferred to Appendix~\ref{apx:baseline_compare}.

\textbf{Evaluation Benchmarks.}
Our main evaluation focuses on mathematical reasoning benchmarks widely used in prior research \citep{bi2024forest,li2025system,lin2025cppo}: MATH500 \citep{lightman2023let}, Minerva-Math \citep{lewkowycz2022solving}, OlympiadBench \citep{he2024olympiadbench}, AMC2023 \citep{amc}, and AIME2024 \citep{aime}. To assess generalizability, we extend our evaluation to BoardgameQA (BGQA) \citep{kazemi2023boardgameqa}, CRUXEval (CRUX) \citep{gu2024cruxeval}, StrategyQA (STGQA) \citep{geva2021did}, TableBench \citep{wu2025tablebench}, and STEM subsets of MMLU-Pro \citep{wang2024mmlu}. Furthermore, we test code generation and language understanding capabilities using HumanEval \citep{chen2021evaluating}, LiveCodeBench \citep{jain2024livecodebench}, and RACE \citep{lai2017race}.

\subsection{Main Results}
\begin{table*}[b] 
    \centering
    \vspace{-10pt}
    \renewcommand{\arraystretch}{1.25}
    \captionsetup{font=small}
    \caption{\textbf{Comprehensive performance comparison.} \projectname~ consistently outperforms trajectory-level optimization baselines across both in-domain mathematical reasoning and out-of-domain generalization tasks.}
    \resizebox{\linewidth}{!}{
    \begin{tabular}{ l cccccc cccccc }
    \toprule
    \multirow{2.5}{*}{\textbf{Models}} & \multicolumn{6}{c}{\textbf{In-Domain Math Reasoning}} & \multicolumn{6}{c}{\textbf{Out-of-Domain Reasoning}} \\
    \cmidrule(lr){2-7} \cmidrule(l){8-13}
    & \makecell{Math \\ 500} & \makecell{Minerva \\ Math} & \makecell{Olympiad \\ Bench} & \makecell{AIME24 \\ \footnotesize(Avg@1/32)} & \makecell{AMC \\ 23} & \textbf{Avg} & BGQA & CRUX & STGQA & TableBench & \makecell{MMLU \\ STEM} & \textbf{Avg} \\
    \midrule
    \textbf{Qwen2.5-Math-7B}  & 64.8 & 15.4 & 25.6 & 16.7 & 37.5 & 32.0 & 48.0 & 50.0 & 88.0 & 38.0 & 40.0 & 52.8 \\
    \quad + GRPO  & 76.2 & 32.7 & 38.1 & 16.7 & 55.0  & 43.7 & 50.5 & 53.0 & 89.5 & 39.0 & 42.0 & 54.8 \\
    \quad + Simple-RL-Zero & 78.0 & 33.1 & 36.6 & 20.0 & 57.5  & 45.0 & 51.5 & 53.5 & 90.0 & 40.0 & 42.5 & 55.5 \\
    \quad + PURE-VR & 79.8 & 36.8 & 41.9 & 20.0 & 57.5 & 47.5 & 52.0 & 54.0 & 90.5 & 40.5 & 43.0 & 56.0 \\
    \quad + DPO-VP & 74.8 & 35.3 & 36.9 & 23.3 & 60.0  & 46.1 & 52.5 & 54.5 & 91.0 & 41.0 & 43.5 & 56.5 \\
    \rowcolor[RGB]{234, 238, 234}
    \quad + \projectname & 76.6 & 37.8 & 43.8 & 23.3 & 62.5  & 48.8 & 53.5 & 55.5 & 91.5 & 41.5 & 44.0 & 57.2 \\
    \rowcolor[RGB]{234, 238, 234}
    \quad + \projectname + CaT & \textbf{81.9} & \textbf{40.5} & \textbf{46.9} & \textbf{26.7} & \textbf{67.5} & \textbf{52.7} & \textbf{56.2} & \textbf{58.3} & \textbf{96.1} & \textbf{43.6} & \textbf{46.2} & \textbf{60.1} \\
    \midrule
    \textbf{Qwen2.5-7B-Instruct}  & 76.2 & 37.6 & 43.0 & 13.3 & 52.5  & 44.4 & 53.0 & 58.1 & 91.3 & 43.2 & 45.2 & 58.2 \\
    \quad + GRPO  & 79.0 & 41.0 & 46.5 & 13.3 & 55.0 & 46.6 & 54.5 & 59.9 & 92.1 & 44.0 & 46.2 & 59.3 \\
    \quad + Simple-RL-Zero & 80.2 & 41.5 & 45.8 & 16.7 & 57.5  & 47.8 & 55.5 & 60.9 & 92.5 & 44.4 & 46.7 & 60.0 \\
    \quad + PURE-VR & 81.5 & 43.0 & 47.5 & 16.7 & 57.5  & 48.9 & 56.0 & 61.4 & 92.8 & 44.7 & 47.0 & 60.4 \\
    \quad + DPO-VP & 79.8 & 42.5 & 46.2 & 20.0 & 60.0 & 49.1 & 56.8 & 62.1 & 93.3 & 45.2 & 47.4 & 61.0 \\
    \rowcolor[RGB]{234, 238, 234}
    \quad + \projectname  & 82.0 & 44.0 & 48.3 & 20.0 & 62.5 & 50.6 & 57.5 & 62.9 & 93.8 & 45.7 & 48.0 & 61.6 \\
    \rowcolor[RGB]{234, 238, 234}
    \quad + \projectname + CaT & \textbf{87.7} & \textbf{47.1} & \textbf{51.7} & \textbf{26.7} & \textbf{67.5} & \textbf{56.1} & \textbf{60.4} & \textbf{66.0} & \textbf{98.5} & \textbf{48.0} & \textbf{50.4} & \textbf{64.7} \\
    \midrule
    \textbf{Llama-3.1-8B-Instruct} & 49.6 & 13.2 & 23.5 & 6.7 & 27.5 & 24.1 & 40.0 & 45.0 & 82.0 & 35.0 & 36.0 & 47.6 \\
    \quad + GRPO  & 52.0 & 15.0 & 25.0 & 6.7 & 30.0  & 25.5 & 41.5 & 46.5 & 83.0 & 35.8 & 37.0 & 48.8 \\
    \quad + Simple-RL-Zero & 53.2 & 15.5 & 25.6 & 10.0 & 30.0  & 26.9 & 42.0 & 47.0 & 83.5 & 36.2 & 37.2 & 49.2 \\
    \quad + PURE-VR & 54.0 & 16.0 & 26.8 & 10.0 & 32.5 & 27.6 & 42.5 & 47.5 & 83.8 & 36.5 & 37.5 & 49.6 \\
    \quad + DPO-VP & 54.8 & 16.5 & 27.2 & 13.3 & 32.5  & 28.8 & 43.0 & 48.0 & 84.2 & 37.0 & 38.0 & 50.0 \\
    \rowcolor[RGB]{234, 238, 234}
    \quad + \projectname  & 55.2 & 15.6 & 27.6 & 13.3 & 35.0 & 29.1 & 43.5 & 48.5 & 84.5 & 37.5 & 38.5 & 50.5 \\
    \rowcolor[RGB]{234, 238, 234}
    \quad + \projectname + CaT & \textbf{59.1} & \textbf{16.7} & \textbf{29.5} & \textbf{20.0} & \textbf{40.0} & \textbf{33.1} & \textbf{45.7} & \textbf{51.0} & \textbf{88.8} & \textbf{39.4} & \textbf{40.4} & \textbf{53.1} \\
    \bottomrule
    \end{tabular}
    }
    \label{tbl:merged_performance}
    \vspace{-10pt}
\end{table*}
\textbf{Training-Based Comparison.} Table~\ref{tbl:merged_performance} presents the comparison between \projectname~ and baseline methods under matched training and inference budgets. \projectname~ delivers consistent gains across all three base models. On Qwen2.5-7B-Instruct, it achieves an average score of 50.6, surpassing GRPO, Simple-RL-Zero, PURE-VR, and DPO-VP. These improvements stem from our \emph{step-wise confidence-aware preference learning}, which aligns token probabilities with intermediate-step correctness more effectively than trajectory-level rewards. The monotonic accuracy growth in Appendix Figure~\ref{fig:acc} further corroborates this, signifying stable self improvement as calibration accumulates.

\begin{wraptable}{R}{0.5\textwidth}
    \vspace{-1.5em}
    \centering
    \renewcommand{\arraystretch}{1.2} 
    \caption{\textbf{\small Comparison of inference strategies.} M500 denotes MATH 500. MM denotes Minerva-Math. OB denotes OlympiadBench. A23 denotes AMC2023. A24 denotes AIME2024.}
    \scriptsize
    \setlength{\tabcolsep}{3pt} 
    \begin{tabular*}{\linewidth}{@{\extracolsep{\fill}} l cccccc @{}}
    \toprule
    \textbf{Models} & \makecell{M500} & \makecell{MM} & \makecell{OB} & \makecell{A24} & \makecell{A23} & \textbf{Avg} \\
    \midrule
    \textbf{Qwen-Math-CASPO} & 76.6 & 37.8 & 43.8 & 23.3 & 62.5  & 48.8\\
    \quad + CoT              & 78.2 & 38.6 & 44.7 & 23.3 & 63.8 & 49.7 \\
    \quad + Self-Consistency & 79.6 & 39.3 & 45.6 & 26.7 & 65.0 & 51.2 \\
    \quad + DiPT             & 80.0 & 39.5 & 45.8 & 23.3  & 65.0  & 50.7 \\
    \rowcolor[RGB]{234, 238, 234} 
    \quad + CaT (Ours)       & \textbf{81.9} & \textbf{40.5} & \textbf{46.9} & \textbf{26.7} & \textbf{67.5} & \textbf{52.7} \\
    \midrule 
    \textbf{Qwen-Ins-CASPO}  & 82.0 & 44.0 & 48.3 & 20.0 & 62.5 & 50.6 \\
    \quad + CoT              & 83.6 & 44.9 & 49.3 & 20.0 & 63.8 & 52.3 \\
    \quad + Self-Consistency & 85.3 & 45.8 & 50.2 & 23.3 & 65.0 & 53.9 \\
    \quad + DiPT             & 85.7 & 46.0 & 50.5 & 20.0  & 65.0 & 53.4  \\
    \rowcolor[RGB]{240, 245, 240}
    \quad + CaT (Ours)       & \textbf{87.7} & \textbf{47.1} & \textbf{51.7} & \textbf{26.7} & \textbf{67.5} & \textbf{56.1} \\
    \midrule
    \textbf{Llama-Ins-CASPO} & 55.2 & 15.6 & 27.6 & 13.3 & 35.0 & 29.1 \\
    \quad + CoT              & 56.3 & 15.9 & 28.1 & 13.3 & 36.3 & 30.0 \\
    \quad + Self-Consistency & 57.4 & 16.2 & 28.7 & 16.7 & 37.5 & 31.3 \\
    \quad + DiPT             & 57.7 & 16.3 & 28.8 & 13.3  & 37.5 & 30.7 \\
    \rowcolor[RGB]{240, 245, 240}
    \quad + CaT (Ours)       & \textbf{59.1} & \textbf{16.7} & \textbf{29.5} & \textbf{20.0} & \textbf{40.0} & \textbf{33.1} \\
    \bottomrule
    \end{tabular*}
    \label{tbl:inferencemath}
    \vspace{-2em} 
\end{wraptable}
\textbf{Inference-Time Comparison.} 
Table~\ref{tbl:inferencemath} evaluates various inference strategies applied to \projectname-trained models. All methods utilize an identical sampling budget ($K{=}10$) to ensure fair comparison \citep{zhang2023sac3}. We observe that both Self-Consistency and CaT yield larger performance deltas on \projectname~ models compared to the original instruct-tuned counterparts. This indicates that the calibration learned during training effectively transfers to inference-time search. Specifically, our CaT strategy achieves the highest average performance across all base models while maintaining the fixed sampling budget, validating the efficacy of pruning low-confidence paths.

\begin{wraptable}{R}{0.5\textwidth}
    \vspace{-1em}
    \centering
    \small
    \caption{Scalability on strong Instructed models.}
    \setlength\tabcolsep{4pt}
    \resizebox{\linewidth}{!}{%
    \begin{tabular}{lcccc}
    \toprule[1pt]
    Method & \makecell{Data\\Budget} & \makecell{Math500\\(Pass@1)} & \makecell{Math500\\(Maj@8)} & \makecell{Olympiad\\Bench} \\
    \midrule
    Base & 2.5M (SFT) & 83.6 & 87.1 & 41.6 \\
    + DPO-VP & +8K & 80.9 & 82.1 & 44.0 \\
    + PCPO  & +30K & 81.4 & 83.8 & 44.3 \\
    \rowcolor[RGB]{234, 238, 234}
    + \projectname~ (Ours) & +8K & \textbf{85.1} & \textbf{90.4} & \textbf{49.0} \\
    \bottomrule[1pt]
    \end{tabular}%
    }
    \label{tbl:strong_model}
    \vspace{-1em}
\end{wraptable}
\textbf{Scalability to Strong Instruction-Tuned Models.} 
We investigate whether \projectname~ provides marginal gains for models already optimized through extensive SFT and alignment. Using Qwen2.5-Math-7B-Instruct (trained on 2.5M samples) as a baseline, Table~\ref{tbl:strong_model} shows that \projectname~ yields substantial improvements with only 8K seed samples, elevating MATH500 Pass@1 from 83.6\% to 85.1\% and Maj@8 to 90.4\%, surpassing strong baselines such as DPO-VP and PCPO.
These results position \projectname~ as a complementary stage that corrects confidence miscalibrations after large-scale SFT.

To verify scalability, we evaluate \projectname~ on Qwen3-8B-Base against tree-search baselines. As shown in Table~\ref{tbl:qwen3}, \projectname~ outperforms rStar-Math and Satori on all benchmarks while using zero reward-model data, compared with 3.64M and 240K samples required by the two baselines. The gains on AIME'24 and AIME'25 further show that calibrated intrinsic uncertainty scales effectively to stronger models without external supervision.

\subsection{Generalization and Transferability}
\label{sec:general_results}
 
\textbf{Out-of-Domain Transferability.}
Although trained exclusively on mathematical data, \projectname~ demonstrates robust transfer capabilities to non-mathematical reasoning tasks (Table~\ref{tbl:merged_performance}). It consistently improves performance across diverse benchmarks, including commonsense (STGQA), code (CRUX), tabular reasoning (TableBench), and STEM knowledge (MMLU-Pro STEM). Specifically, on the aggregated MMLU-Pro subsets (spanning physics, chemistry, CS, engineering, biology, and economics; 5,371 problems), \projectname~ improves Qwen2.5-Math-7B from 52.8\% to 57.2\% and Qwen2.5-7B-Instruct from 58.2\% to 61.6\%. These gains indicate that our stepwise aggregation strategy generalizes beyond the math domain, offering a lightweight yet robust mechanism for diverse reasoning.

\begin{figure}[!h]
    \centering
    \small
    \includegraphics[height=4.5cm, keepaspectratio]{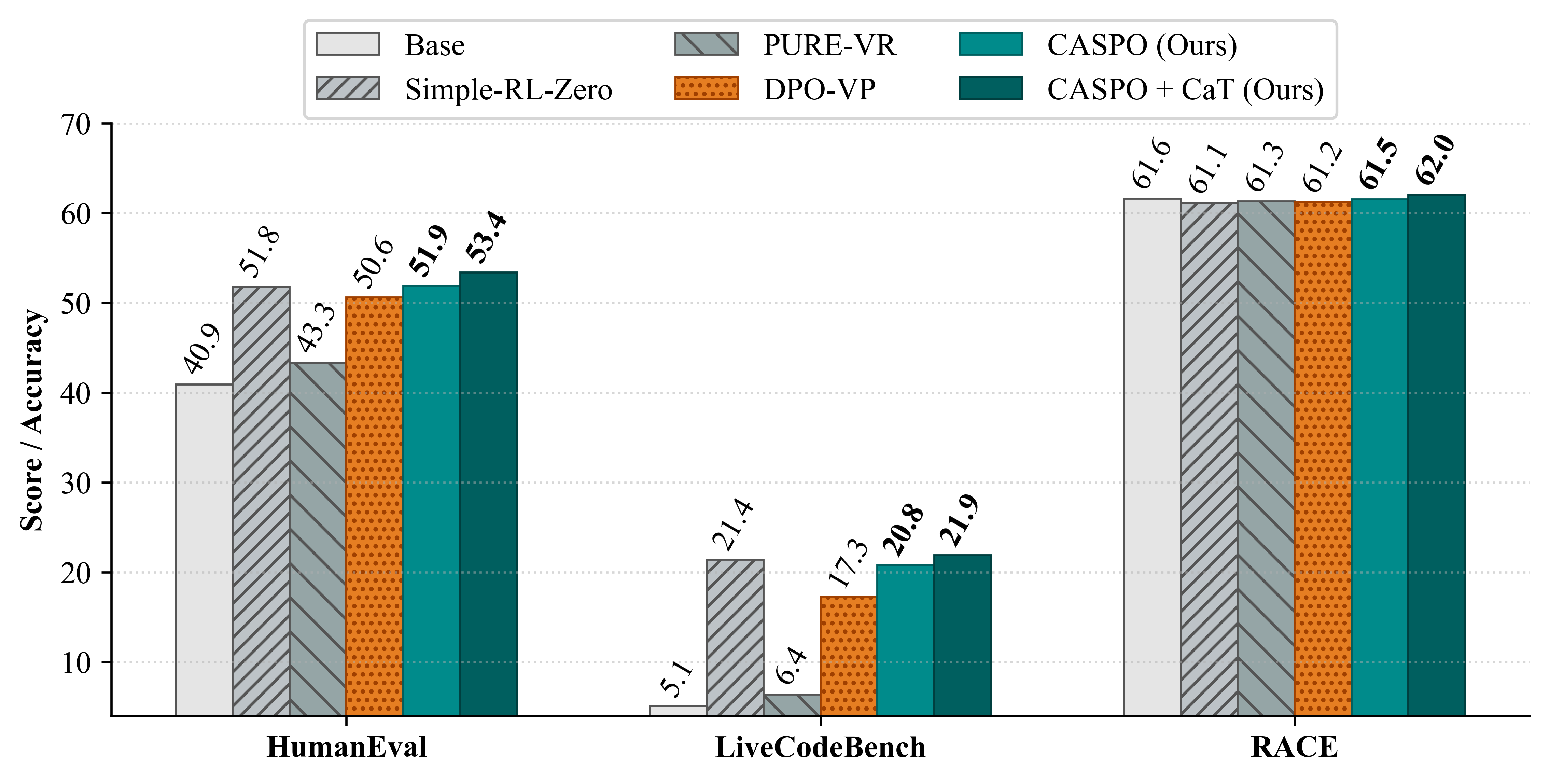}
    \caption{Generalization Performance of \projectname~ on Qwen2.5-Math-7B across HumanEval, LiveCodeBench, and RACE Benchmarks.}
    \label{fig:caspocode}
    \vspace{-1.5em}
\end{figure}

\textbf{Generalization to Code and Language Tasks.} 
To verify that \projectname~ captures general reasoning consistency rather than overfitting to mathematical patterns, we extended our evaluation to strictly non-mathematical domains: code generation (HumanEval \citep{chen2021evaluating}, LiveCodeBench \citep{jain2024livecodebench}) and reading comprehension (RACE \citep{lai2017race}). As detailed in Figure~\ref{fig:caspocode}, \projectname~ consistently outperforms baselines across these diverse tasks. On HumanEval, it improves the base model's Pass@1 from 40.9\% to 51.9\%. These results confirm that identifying and pruning low-confidence steps is a fundamental reasoning capability that transfers effectively across modalities.

\textbf{Computational Efficiency and Signal Complexity.}
A critical advantage of \projectname~ is its computational frugality compared to methods relying on external verifiers or extensive sampling. Prior approaches such as Process Reward Models (PRM) or Process Preference Models (PPM) \citep{guan2025rstar} typically require a full model forward pass to evaluate each intermediate step, resulting in a computational complexity of $O(L^2d)$ where $L$ is sequence length. In contrast, \projectname~ computes the verification signal directly from the output logits already generated by the policy model. As shown in Table~\ref{tbl:complexity}, the complexity of our entropy calculation is $O(V)$ (vocabulary size) and is independent of the sequence length. This reduces the verifier latency by two orders of magnitude (from 2.9s to 0.03s per step), making intrinsic entropy a negligible cost for scalable oversight.
 \begin{wraptable}{r}{0.4\textwidth}
    \centering
    \small
    \vspace{-1em}  
    \caption{Complexity and latency comparison on Qwen2.5-Math-7B.}
    \setlength\tabcolsep{1.4pt}
    \begin{tabular}{l|c|c|c}
    \toprule
    Model & \makecell{Math500 \\ (Acc)} & \makecell{Latency \\ (s/step)} & \makecell{Complexity} \\
    \midrule
    Base & 64.8 & -- & -- \\
    + PRM & 76.3 & 2.9 & $O(L^2d)$ \\
    + PPM \citep{guan2025rstar} & 78.4 & 1.4 & $O(L^2d)$ \\
    \rowcolor[RGB]{234, 238, 234}
    + Ours & \textbf{81.9} & \textbf{0.03} & $\mathbf{O(V)}$ \\
    \bottomrule
    \end{tabular}
    \label{tbl:complexity}
\end{wraptable}
\textbf{Latency Overhead of CaT Inference.}
We further evaluate the inference overhead introduced by CaT. Unlike Self-Consistency, which relies on repeated sampling and multiple full forward passes, CaT uses the token-level entropy already available during generation to guide a small number of candidate continuations. This design focuses computation on uncertain reasoning regions while pruning low-confidence paths early. As shown in Table~\ref{tbl:latency_cat}, CaT achieves higher accuracy with modest additional latency: on Qwen2.5-7B-Instruct, its end-to-end latency is 2.8 s/query, close to greedy decoding at 1.2 s/query and much lower than Self-Consistency at 12.5 s/query.
\begin{table}[!t]
    \centering
    \caption{\small \textbf{Inference latency analysis.} On Qwen2.5-7B-Instruct and Llama-3.1-8B-Instruct, CaT achieves gains with marginal latency overhead over greedy decoding, whereas Self-Consistency is computationally costly.}
    \small
    \begin{tabular}{l|cc|cc}
    \toprule
    & \multicolumn{2}{c|}{Qwen2.5} & \multicolumn{2}{c}{Llama-3.1} \\
    Method & Math & Latency (s/query) & Math & Latency (s/query) \\
    \midrule
    Greedy Decoding & 82.0 & 1.2 & 55.2 & 1.5 \\
    Chain-of-Thought & 83.6 & 4.6 & 56.3 & 5.9 \\
    Self-Consistency & 85.3 & 12.5 & 57.7 & 18.0 \\
    \rowcolor[RGB]{234, 238, 234}
    CaT (Ours) & \textbf{87.7} & \textbf{2.8} & \textbf{59.1} & \textbf{3.1} \\
    \bottomrule
    \end{tabular}
    \label{tbl:latency_cat}
    \vspace{-10pt}
\end{table}

\begin{wraptable}{r}{0.4\linewidth}
    \centering
    \small
    \vspace{-10pt}
    \caption{\small \textbf{Calibration quality on MATH-500.}
    Base denotes Qwen2.5-Math-7B.
    }
    \setlength\tabcolsep{4pt}
    \begin{tabular}{lccc}
    \toprule[1pt]
    Model & Acc. (\%) & ECE $\downarrow$ & BS $\downarrow$ \\
    \midrule
     Base & 64.8 & 0.184 & 0.215 \\
    + DPO & 71.2 & 0.142 & 0.188 \\
    \rowcolor[RGB]{234, 238, 234}
    + \projectname~ & \textbf{76.6} & \textbf{0.081} & \textbf{0.142} \\
    \bottomrule[1pt]
    \end{tabular}
    \label{tbl:calibration}
    \vspace{-10pt}
\end{wraptable}
\textbf{Step-wise correctness rate before/after CASPO.}
To verify that \projectname~ achieves genuine calibration rather than merely improving accuracy, we compute the Expected Calibration Error (ECE) and Brier Score (BS) on MATH-500 (Table~\ref{tbl:calibration}). While standard DPO improves accuracy, it remains poorly calibrated (ECE = 0.142). \projectname~ substantially reduces ECE to 0.081 and Brier Score to 0.142, no external that the model's confidence becomes a more reliable indicator of correctness after alignment. This result directly supports the core motivation of our framework: high predictive confidence should be strictly reserved for valid logical steps.

 \begin{wraptable}{r}{0.4\linewidth}
    \centering
    \small
    \vspace{-10pt}  
    \caption{Step Correctness AUC-ROC.}
    \setlength\tabcolsep{4pt}
    \begin{tabular}{lc}
    \toprule[1pt]
    Uncertainty Signal & AUC-ROC \\
    \midrule
    Continuation Length & 0.54 \\
    Max Token Probability & 0.68 \\
    Perplexity (PPL) & 0.72 \\
    \rowcolor[RGB]{234, 238, 234}
    Shannon Entropy (Ours) & \textbf{0.86} \\
    \bottomrule[1pt]
    \end{tabular}
    \label{tbl:aucroc}
    \vspace{-10pt}  
\end{wraptable}
\textbf{Entropy as a Step-Correctness Signal.}
To justify the use of Shannon entropy over alternative uncertainty signals, we compare their predictive power for step-wise correctness via AUC-ROC (Table~\ref{tbl:aucroc}). Shannon entropy achieves an AUC-ROC of 0.86, outperforming continuation length (0.54), max token probability (0.68), and perplexity (0.72). The key distinction is that perplexity reflects the probability of the chosen token sequence, whereas entropy measures the competitiveness of the entire vocabulary distribution, capturing the model's confusion between diverging logical paths even when the top-1 token has high probability. Continuation length is a noisier post-hoc signal that conflates rigorous derivations with hallucination loops.

\begin{figure*}[h]
    \centering
    \small
    \includegraphics[width=\linewidth]{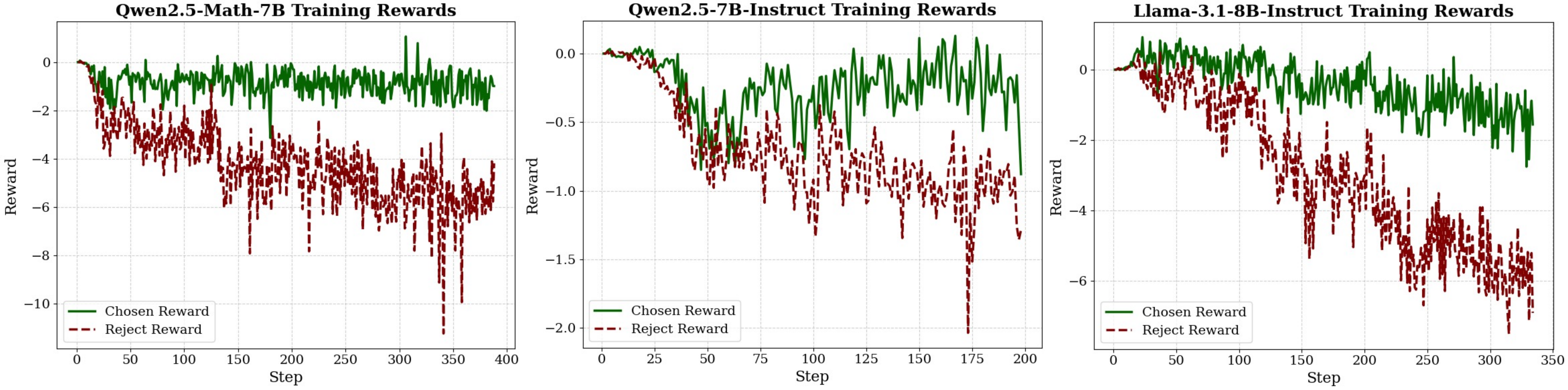}
    \caption{\textbf{Training dynamics.} Reward evolution during DPO training across Qwen2.5-Math-7B, Qwen2.5-7B-Instruct, and Llama-3.1-8B-Instruct models.}
    \label{fig:reward}
    \vspace{-10pt}
\end{figure*}

\begin{wraptable}{R}{0.4\textwidth} 
    \centering
    \small
    \renewcommand{\arraystretch}{0.9} 
    \caption{Token-level entropy gap.
    C.S. denotes Correct Step. I.S. denotes Incorrect Step. E.G. denotes Entropy Gap.}
    \setlength\tabcolsep{5pt} 
    \begin{tabular}{lccc}
    \toprule[1pt]
    \textbf{Stage} & \makecell{\textbf{C.S.} \\ } & \makecell{\textbf{I.S.} \\ } & \textbf{E.G.} \\
    \midrule
    Qwen2.5-7B & 0.38 & 0.42 & 0.04 \\
    \rowcolor[RGB]{234, 238, 234}
    + \projectname~ & \textbf{0.22} & \textbf{0.88} & \textbf{0.66} \\
    \bottomrule[1pt]
    \end{tabular}
    \label{tbl:entropy_gap}
\end{wraptable}
Table~\ref{tbl:entropy_gap} further illustrates why entropy is an effective pruning signal. Before training, the model is often confidently wrong (entropy gap between correct and incorrect steps = 0.04). After \projectname~, incorrect steps exhibit a sharp entropy increase (gap = 0.66), providing a clear and reliable signal for CaT to prune logically flawed branches.
\textbf{Training Dynamics.}
Following prior work on DPO training dynamics \citep{ren2024learning}, we examine reward evolution during optimization. As shown in Figure~\ref{fig:reward}, all models exhibit expected dual-pressure mechanism: chosen rewards initially drop before recovering near zero, while rejected rewards decrease monotonically, confirming theoretical framework of simultaneous upward and downward pressures. For completeness, we examine the evolution of training accuracy and loss, with corresponding curves provided in Appendix~\ref{apx:acc_loss}.

Our results reveal clear model-specific patterns: Qwen2.5-Math-7B converges the fastest and with the greatest stability, achieving the largest reward separation of about 6.0. This large reward separation reflects its strong alignment with mathematical reasoning preferences, strengthened by domain-specific pre-training. Qwen2.5-7B-Instruct converges efficiently within 200 steps, reaching a moderate separation of about 1.5, which indicates a balance between training efficiency and preference learning. In contrast, Llama-3.1-8B-Instruct shows higher volatility during optimization but achieves a separation comparable to the Math model, although this requires more careful tuning of hyperparameters.

\begin{figure*}[b]
    \centering
    \small
    \includegraphics[width=\linewidth]{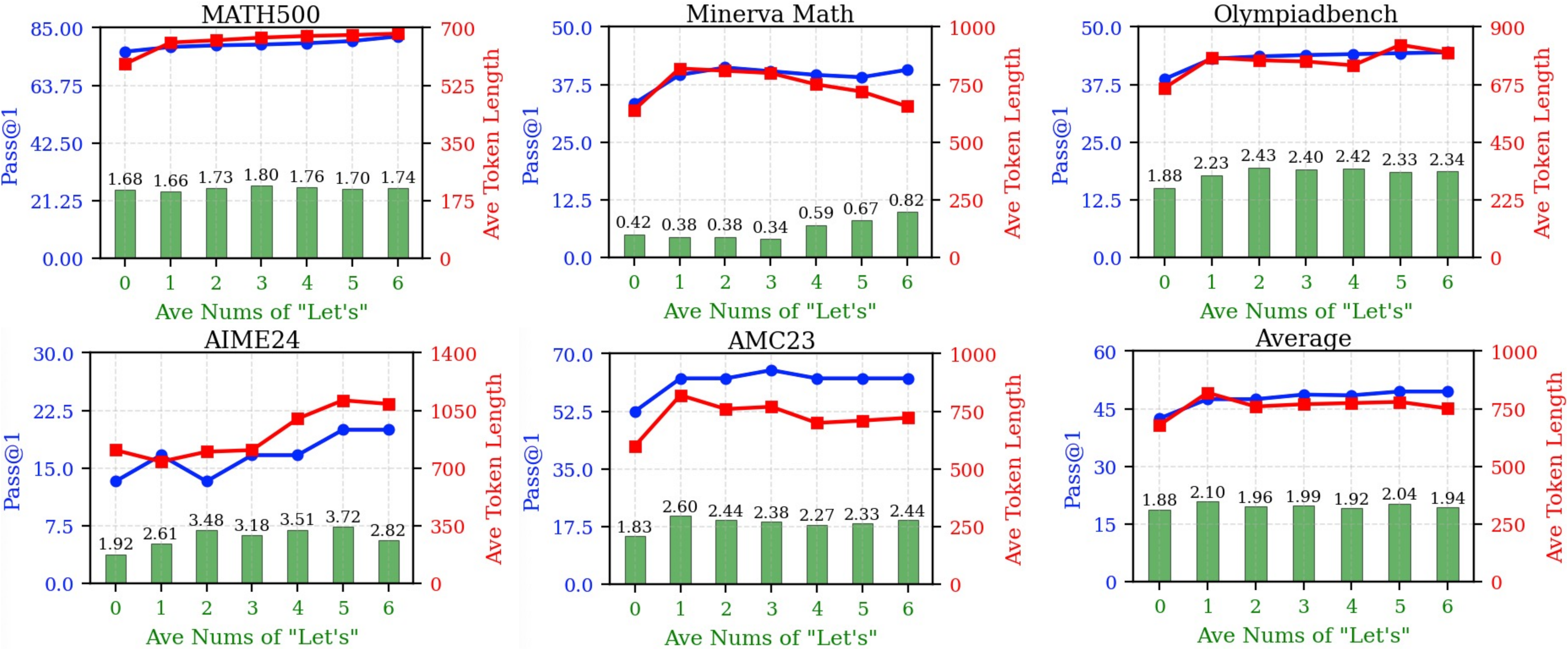}
    \caption{\textbf{Evolution of token length and self-correction.} Pass@1 accuracy improves consistently across DPO rounds without substantial increase in token length. Meanwhile, the use of self-talk triggers declines or stabilizes, suggesting that DPO guides models toward more concise reasoning.}
    \label{fig:length}
    \vspace{-1em}
\end{figure*} 
 
\paragraph{Token Length and Reasoning Pattern Evolution.}
To examine whether the observed performance improvements stem merely from generating longer reasoning chains, we analyze both the token length and reasoning patterns of Qwen2.5-7B-Instruct. As shown in Figure~\ref{fig:length}, Pass@1 accuracy improves consistently across DPO rounds while the average token length remains stable. Furthermore, we use the frequency of the self-correction trigger ``Wait'' or ``Let's'' as a proxy for explicit self-checking \citep{tu2025enhancing, zhou2025r1}. The observed decline in these triggers suggests that \projectname~ does not teach the model to mimic reflective phrasing. Instead, it internalizes the verification process. The model learns to rely on the optimized preference signals to produce correct answers directly.

\subsection{Discussion}
\paragraph{Decoupling Confidence from External Supervision.}
We decouple the model's intrinsic logical confidence from the role of external supervision. In \projectname~, the external evaluator serves only to verify final answer correctness during the offline data collection phase. This procedure mirrors the established paradigm in mathematical reasoning research, where datasets such as GSM8K \citep{cobbe2021training} or MATH500 \citep{lightman2023let} utilize automated ground truth verification to filter training trajectories. This one-time investment during dataset construction ensures that the model requires no external guidance during deployment. More importantly, the core learning signal in our framework originates from the model's own token-level entropy rather than the evaluator's feedback. By extracting correctness and confidence from these two independent channels, we ensure that the supervision remains stable even if the evaluator occasionally mislabels a reasoning path. The confidence-aware signal acts as an internal anchor that prioritizes stable, mastered reasoning over accidental success. Ultimately, this separation allows the model to internalize the verification process, enabling efficient and autonomous inference without the computational burden of an external judge.
 
\subsection{Ablations}
\label{sec:ablations}
\textbf{Iterative Training.}
We evaluate iterative training by applying \projectname~ over three epochs, where training data is regenerated by the current policy at each stage. Unlike standard fine-tuning on static datasets, this allows supervision to track the model's evolving reasoning distribution. Figure~\ref{fig:epoch} shows monotonic improvements: the first epoch yields the largest gains, with Math500 improving from 64.8\% to 76.6\% and AMC23 increasing from 37.5\% to 60.0\%. This surge indicates the rapid rectification of primary calibration errors. Subsequent iterations induce more granular refinements, pushing AMC23 further to 62.5\% and OlympiadBench from 37.8\% to 38.7\%. This reflects progressive optimization where early stages establish a confidence baseline while later ones refine boundary handling, validating a positive feedback loop where superior policies generate higher-fidelity supervision.

\textbf{Balance between Diversity and Reliability.}
Results in Table~\ref{tbl:inferencemath} show that \projectname~ improves the balance between diversity and reliability by providing more accurate confidence signals. This makes pass@$k$ sampling less noisy and better aligned with the model's calibrated preferences. Aggregation methods such as majority voting or Self-Consistency further benefit from these higher-quality candidates, which in turn explains why CaT achieves stronger and more stable gains.
\section{Conclusion}
\label{sec.conclusion}

This work addresses the critical discrepancy between final answer accuracy and the logical integrity of intermediate reasoning steps. We demonstrate that reliance on external verifiers or exhaustive sampling is not the only path to reliable reasoning. By introducing ~\projectname{}, we show that a model's intrinsic token level uncertainty provides a powerful and efficient signal for alignment. Building on this insight, our framework bridges the gap between training and inference, using confidence aware preference optimization to calibrate the model and the CaT strategy to dynamically refine reasoning trajectories with minimal latency. Experimental results confirm that ~\projectname{} fundamentally enhances the faithfulness of the reasoning process rather than merely inflating benchmark scores. By leveraging intrinsic uncertainty, the model learns to identify and correct logical inconsistencies without heavy external supervision. As a result, ~\projectname{} enables a scalable and transparent framework for improving reasoning reliability. Our released dataset and analysis support future work on fine-grained, step-wise alignment and the diagnosis of hidden reasoning failures.

\newpage

\bibliographystyle{plain}
\bibliography{main}

\clearpage
\appendix
\section*{Limitations}
Although \projectname~ achieves consistent gains across different benchmarks and model families, there are several limitations worth noting. First, our definition of confidence is based on Shannon entropy. This provides a simple and effective way to reduce logical hallucinations and improve calibration, but it is only one possible choice among many uncertainty measures. Other signals, such as model self-reflection or internal representations, may capture different aspects of uncertainty. A more systematic comparison of these alternatives would be a useful direction for future work. Second, our data construction pipeline relies on an offline step-wise evaluator, which may introduce evaluator-specific biases when the evaluator shares similar reasoning patterns with the target model. Although the evaluator is used only during offline data construction, future work could explore self-contained or jointly trained verification mechanisms to make the framework more robust and scalable.

\section*{Broader Impact}
\projectname~ encourages models to optimize reasoning step by step and to better align confidence with correctness. In doing so, it can improve the reliability and transparency of language-model reasoning, especially in tasks where intermediate steps matter. However, stronger reasoning ability can also increase risks in high-stakes or dual-use scenarios. For example, models may produce more convincing outputs even when they are wrong or are used for harmful purposes. We therefore emphasize the need for careful evaluation before deployment, particularly in downstream applications where errors may have serious consequences, as well as appropriate safeguards against misuse.

We provide additional experimental details, supplementary results, and implementation analysis in the appendix. Specifically, Appendix~\ref{apx:experi} details the training setup, evaluation protocol, baselines, and benchmarks, while Appendix~\ref{apx:results} reports scalability results, training dynamics, aggregation-function ablations, and the full \projectname{} algorithm.

\section{Experimental Setup}
\label{apx:experi}

\subsection{Details of Training and Evaluation}
\label{sec:training-details}

The base models include Qwen2.5-Math-7B, Qwen2.5-7B-Instruct, and Llama-3.1-8B-Instruct. All model-centric training was conducted with full-parameter fine-tuning using the Open-RLHF framework \cite{hu2024openrlhf}. Random seeds are fixed at 42 for reproducibility. All experiments are trained on 4 NVIDIA A800 GPUs (80GB) with mixed-precision (FP16) enabled.

\paragraph{Optimization hyperparameters.}
The SFT stage uses a learning rate of $5\times10^{-6}$, while the Direct Preference Optimization (DPO) stage adopts $5\times10^{-7}$ to stabilize preference-based updates. Both stages share a maximum sequence length of 2048 tokens and a batch size of 64. The DPO loss coefficient $\beta$ is fixed at 0.1. For each DPO round, candidate responses were sampled with temperature $t=0.7$, and preference pairs were filtered according to verifiable-pair criterion in Section~\ref{sec:catot}.

\paragraph{Training schedule.}
Each training run lasts six epochs. For the first three epochs, the sampling temperature is fixed at $t=0.7$ to keep the data distribution close to the initial policy. For epochs four and five, it is increased to $t=1.0$, and further raised to $t=1.2$ in the final epoch. This annealed schedule reflects the observation that performance plateaus after three epochs, while higher temperatures promote exploration of novel reasoning paths without destabilizing optimization.

\paragraph{Evaluation protocol.}
We follow the Qwen-Math evaluation suite. For every benchmark and model, generations are produced with greedy decoding ($t{=}0.0$), one output per input (no sampling, no self-consistency), and a 2048-token generation limit. All models use the same zero-shot CoT prompt template (shown below) to avoid prompt-engineering confounds. We report pass@1. For datasets that provide official scoring scripts, we use those scripts; otherwise, answers are extracted from the boxed span (see below) and matched after standard normalization.

\subsection{Details of Baselines}
\label{apx:baseline_compare}

\paragraph{Training-based methods.}
We compare \projectname~ with representative training-based self-improvement methods that update model parameters using verifiable feedback. GRPO~\citep{shao2024deepseekmath} and Simple-RL-Zero~\citep{zeng2025simplerl} perform on-policy reinforcement learning with verifiable rewards. PURE-VR~\citep{cheng2025stop} propagates verifiable rewards across reasoning steps, while DPO-VP~\citep{tu2025enhancing} applies iterative DPO to verifiable correct--incorrect output pairs. These methods improve reasoning performance, but mainly optimize complete trajectories or final-answer preferences rather than explicitly calibrating step-wise confidence.

\paragraph{Inference-time methods.}
We also compare with inference-time methods that modify decoding without updating model parameters. Chain-of-Thought prompting~\citep{kojima2022large} elicits intermediate reasoning steps, Self-Consistency~\citep{wang2022self} aggregates multiple sampled chains by majority voting, and DiPT~\citep{just2024dipt} uses diverse prompts to improve reasoning coverage. These methods enhance robustness but do not explicitly calibrate confidence or prune unreliable reasoning paths.

\paragraph{Distinction from prior methods.}
Unlike the above methods, \projectname~ aligns confidence with correctness at the reasoning-step level. During training, it uses correct-but-uncertain and confidently incorrect steps to construct preference pairs; during inference, CaT expands or prunes trajectories according to cumulative step-wise confidence. This unified design enables more reliable supervision and more efficient search.

\subsection{Details of Benchmarks}
\label{apx:dataset}

\textbf{MATH500}~\citep{lightman2023let} is a 500-problem subset of the MATH benchmark~\citep{hendrycks2021measuring}. It is uniformly sampled across subjects and difficulty levels, making it used for evaluating mathematical reasoning.

\textbf{Minerva-Math}~\citep{lewkowycz2022solving} consists of 272 challenging mathematical problems. Some questions also involve scientific reasoning in related domains such as physics.

\textbf{OlympiadBench}~\citep{he2024olympiadbench} is a bilingual benchmark containing 8,476 Olympiad-level mathematics and physics problems, including problems adapted from the Chinese college entrance examination. We use its text-only, open-ended mathematics competition subset, which contains 674 problems.

\textbf{AMC2023} and \textbf{AIME2024} are competition-style mathematical reasoning benchmarks. AMC2023 contains 40 text-only problems from the 2023 American Mathematics Competition, while AIME2024 contains 30 text-only problems from the 2024 American Invitational Mathematics Examination.

\textbf{BoardgameQA (BGQA)}~\citep{kazemi2023boardgameqa} is a logical reasoning dataset with 15K unique problems designed to evaluate LLMs' ability to perform defeasible reasoning, where contradictions must be resolved using credibility or recency cues.

\textbf{CRUXEval}~\citep{gu2024cruxeval} evaluates code reasoning and execution. It contains 800 short Python functions, each paired with input-output examples, where models are required to predict the correct output given a function snippet and input.

\textbf{StrategyQA}~\citep{geva2021did} contains 2,780 multi-hop reasoning questions whose reasoning steps are implicit and must be inferred. Each example is paired with a decomposition into sub-steps and supporting evidence from Wikipedia.

\textbf{TableBench}~\citep{wu2025tablebench} evaluates tabular reasoning in real-world data analysis tasks across 18 domains. We use the fact-checking and numerical reasoning subsets, resulting in 491 unique problems that cover fact verification, numerical calculation, and reasoning over structured tables.

\textbf{MMLUPro-STEM}~\citep{wang2024mmlu} is a STEM-focused subset of MMLU-Pro, an enhanced version of MMLU~\citep{hendrycks2020measuring} with more reasoning-intensive questions and expanded answer choices. We select six STEM domains, physics, chemistry, computer science, engineering, biology, and economics, and exclude mathematics to avoid overlap with in-domain mathematical reasoning benchmarks.

\section{Additional Results}
\label{apx:results}

\subsection{Scalability to Stronger Base Models}

As shown in Table~\ref{tbl:qwen3}, \projectname~ achieves the best results across MATH500, AIME'24, and AIME'25 while using zero reward-model data. In comparison, rStar-Math and Satori rely on 3.64M and 240K reward-model samples, respectively. The gains are especially clear on the more challenging AIME benchmarks, where \projectname~ improves AIME'24 to 36.7 and AIME'25 to 33.3. These results indicate that calibrated intrinsic confidence can serve as an efficient alternative to external reward-model supervision and remains effective on stronger base models.

\begin{table}[b]
    \centering
    \scriptsize
    \caption{\textbf{Scalability to Qwen3 and comparison with tree-search baselines.} \projectname~ achieves superior performance on Qwen3-8B-Base using zero reward model data, outperforming rStar-Math \citep{guan2025rstar} (3.64M RM samples) and Satori \citep{shen2025satori} (240K RM samples).}
    \setlength\tabcolsep{3.5pt}
    \begin{tabular}{lcccc}
    \toprule[1pt]
    Method & RM Data & \makecell{MATH500\\(Pass@1)} & AIME'24 & AIME'25 \\
    \midrule
    Qwen3-8B-Base & -- & 87.4 & 23.3 & 20.0 \\
    + rStar-Math & 3.64M & 88.2 & 30.0 & 23.3 \\
    + Satori & 240K & 88.6 & 30.0 & 26.7 \\
    \rowcolor[RGB]{234, 238, 234}
    + \projectname~ (Ours) & 0 & \textbf{89.0} & \textbf{36.7} & \textbf{33.3} \\
    \bottomrule[1pt]
    \end{tabular}
    \label{tbl:qwen3}
    \vspace{-10pt}
\end{table}

\subsection{Effect of Iterative Training}

\begin{figure}[htbp]
  \centering
  \vspace{-10pt}
  \small
  \includegraphics[width=0.98\linewidth]{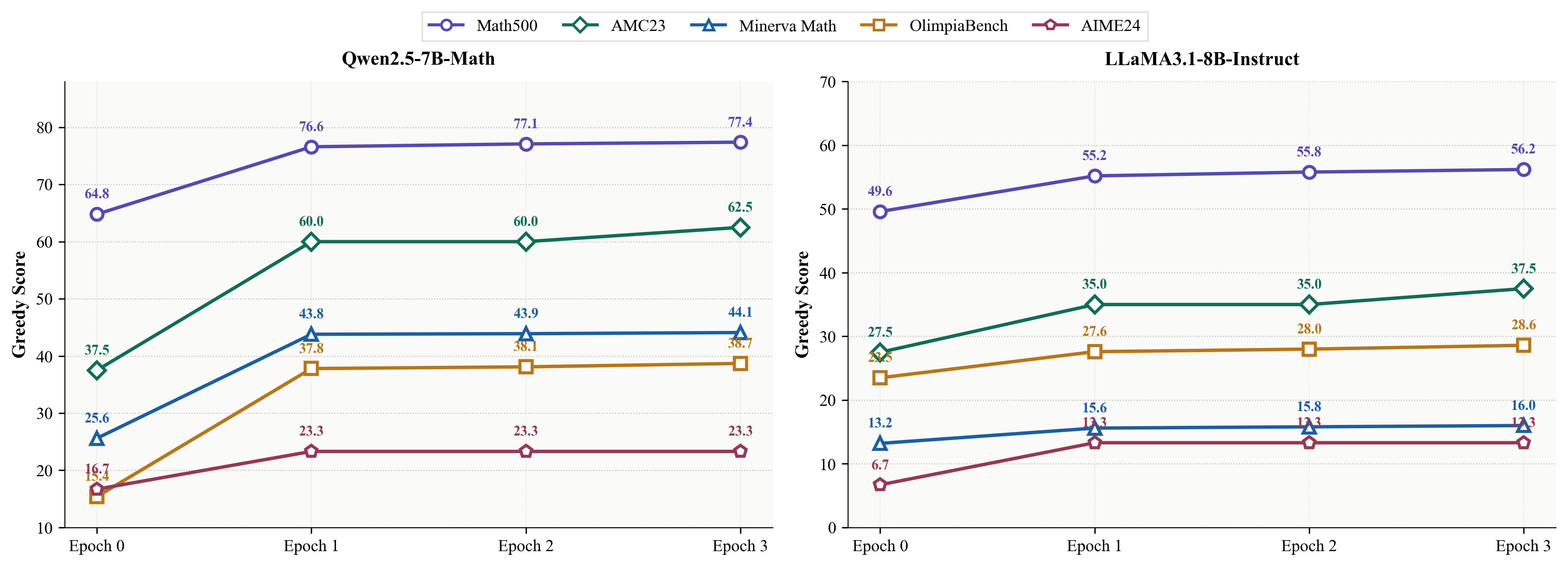}
  \caption{Greedy evaluation scores across iterative \projectname{} training epochs on Qwen2.5-7B-Math (left) and LLaMA3.1-8B-Instruct (right). Both models achieve the largest gains in the first epoch and continue to improve gradually in later epochs.}
  \label{fig:epoch}
  \vspace{-10pt}
\end{figure}

Figure~\ref{fig:epoch} shows greedy evaluation scores across training epochs under iterative \projectname{} training. Both Qwen2.5-7B-Math and LLaMA3.1-8B-Instruct achieve the largest gains in the first epoch, indicating that the initial round corrects major confidence miscalibrations. Subsequent epochs bring smaller but consistent improvements on most benchmarks, suggesting that iterative data regeneration helps refine harder reasoning cases and progressively improves confidence calibration.

\subsection{Accuracy and Loss Dynamics.}
\label{apx:acc_loss}



\begin{figure*}[b]
    \centering
    \small
    \subfloat[Training accuracy trajectories.]{
        \includegraphics[width=0.9\linewidth]{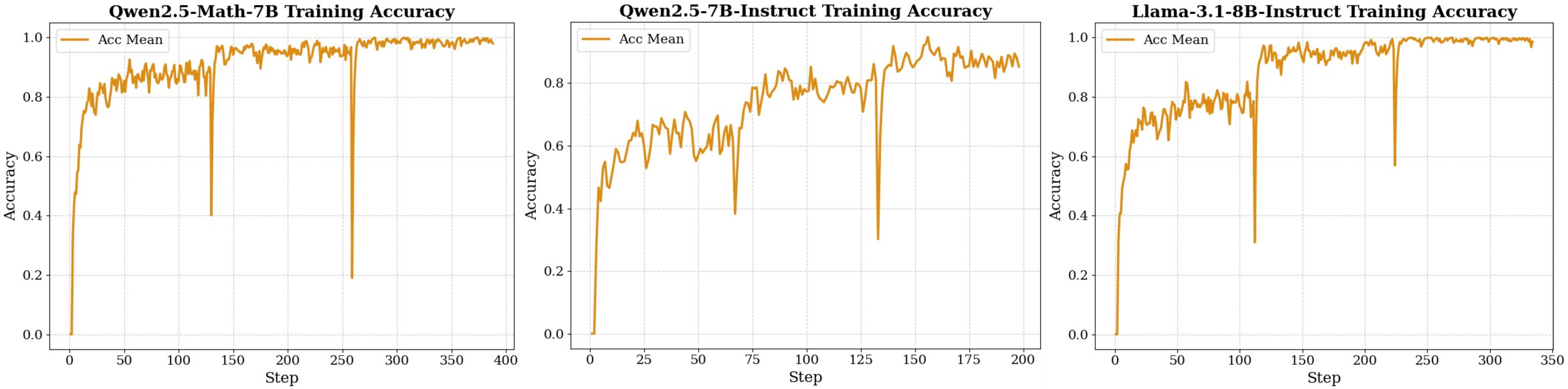}
        \label{fig:acc}
    }

    \vspace{2pt}

    \subfloat[(Loss reduction patterns.]{
        \includegraphics[width=0.9\linewidth]{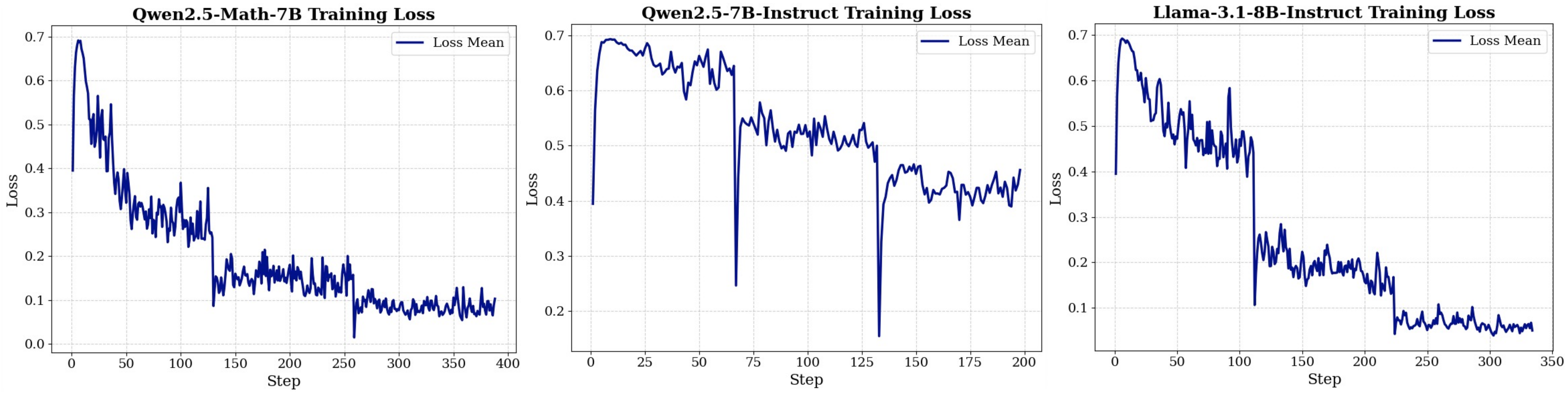}
        \label{fig:loss}
    }
    \caption{Training dynamics during DPO optimization across Qwen2.5-Math-7B, Qwen2.5-7B-Instruct, and Llama-3.1-8B-Instruct.}
    \vspace{-10pt}
    \label{fig:acc_loss}
\end{figure*}

Figures~\ref{fig:acc} and \ref{fig:loss} show the accuracy and loss dynamics during training. Across all models, accuracy increases as reward separation emerges, while loss decreases steadily, indicating that preference learning improves both reward discrimination and prediction reliability. Qwen2.5-Math-7B shows the smoothest convergence, with accuracy quickly approaching high levels and loss declining consistently. Qwen2.5-7B-Instruct stabilizes within about 200 steps, while Llama-3.1-8B-Instruct converges more slowly with larger loss fluctuations but still reaches strong final accuracy.

\subsection{Step-wise Aggregation Function}
\label{apx:agg_results}

We study two choices for aggregating token-level confidence into a step-wise score. Mean entropy measures the model's average uncertainty over the generated tokens:
\begin{equation}
    f_{\text{entropy}}(s) = -\frac{1}{n}\sum_{i=1}^{n} \sum_{v \in \mathcal{V}} p(t_i = v) \log p(t_i = v).
\end{equation}
Multiplicative probability estimates the likelihood of the whole step:
\begin{equation}
    f_{\text{mult}}(s) = \prod_{i=1}^{n} p(t_i).
\end{equation}

Table~\ref{tab:agg_results} compares two choices: Mean entropy better captures uncertainty calibration, while multiplicative probability favors high-likelihood steps and penalizes any low-confidence token. Results show that both signals are useful, but entropy provides stronger overall performance.

\begin{table*}[t]
\centering
\small
\setlength{\tabcolsep}{8pt}
\renewcommand{\arraystretch}{0.86}
\begin{tabular}{lcccc}
\toprule
\textbf{Model} 
& \multicolumn{3}{c}{\textbf{Math}} 
& \textbf{Open-domain} \\
\cmidrule(lr){2-4} \cmidrule(l){5-5}
& \textbf{Math500} 
& \textbf{Minerva Math} 
& \textbf{OlympiadBench} 
& \textbf{MMLU-STEM} \\
\midrule

\multicolumn{5}{l}{\textbf{Multiplication}} \\
Qwen2.5-Math-7B        & 63.2 & 14.7 & 24.9 & 41.9 \\
Qwen2.5-7B-Instruct    & 80.5 & 32.7 & 38.1 & 45.2 \\
Llama3.1-8B-Instruct   & 48.7 & 12.8 & 22.6 & 36.1 \\

\addlinespace[2pt]
\multicolumn{5}{l}{\textbf{Entropy}} \\
Qwen2.5-Math-7B        & 64.8 & 15.4 & 25.6 & 42.5 \\
Qwen2.5-7B-Instruct    & 83.2 & 33.5 & 38.4 & 45.6 \\
Llama3.1-8B-Instruct   & 49.6 & 13.6 & 23.5 & 36.0 \\
\bottomrule
\end{tabular}
\caption{Accuracy of LRMs using multiplicative and entropy aggregation.}
\label{tab:agg_results}
\vspace{-8pt}
\end{table*}

\subsection{CASPO Training Procedure}
\label{apx:caspo_algorithm}

Algorithm~\ref{alg:caspo_compact} summarizes the training procedure of CASPO. It first constructs step-wise preference pairs by comparing the model's confidence with the correctness signal from an offline critic. Correct but low-confidence steps are treated as preferred over competing alternatives, while incorrect steps are paired against the critic-provided correct step. 

\begin{algorithm}[h]
\caption{CASPO Training}
\small
\label{alg:caspo_compact}
\begin{algorithmic}[1]
   \STATE {\bfseries Input:} Math dataset $\mathcal{D}_\text{math}$, target model $\pi_\theta$, critic $\pi_\text{critic}$, confidence threshold $\tau$, iterations $K$.
   \STATE {\bfseries Initialize:} Preference dataset $\mathcal{D} \gets \{\}$, reference model $\pi_\text{ref} \gets \pi_\theta$.

   \FOR{each question $x \in \mathcal{D}_\text{math}$}
       \FOR{each sub-question $q_j$ of $x$}
           \STATE Generate answer $s_j \sim \pi_\theta(\cdot|q_j)$ and confidence $c_j \gets \mathrm{confidence}(s_j|q_j)$.
           \STATE Obtain reference step $g_j \gets \pi_\text{critic}(\cdot|q_j)$.
           \STATE Set $(y_w, y_l) \gets
               \begin{cases}
                   (s_j, \text{competing candidate}) & \text{if } s_j=g_j \text{ and } c_j \le \tau,\\
                   (g_j, s_j) & \text{if } s_j \ne g_j,\\
                   \text{skip} & \text{otherwise.}
               \end{cases}$
           \STATE Add $(q_j, y_w, y_l)$ to $\mathcal{D}$ if not skipped.
       \ENDFOR
   \ENDFOR

   \FOR{$k = 1$ {\bfseries to} $K$}
       \STATE $\pi_{\theta_k} \gets \arg\min_\theta \mathcal{L}_{\mathrm{DPO}}(\pi_\theta, \pi_\text{ref}; \mathcal{D})$.
       \STATE $\pi_\text{ref} \gets \pi_{\theta_k}$.
   \ENDFOR

   \STATE {\bfseries Return:} Optimized model $\pi_{\theta_K}$.
\end{algorithmic}
\end{algorithm}

\subsection{CaT Inference Procedure}
\label{apx:cat_algorithm}

Algorithm~\ref{alg:cat_inference} summarizes the inference-time procedure of CaT. CaT does not impose a predefined reasoning format; steps are segmented by natural delimiters such as line breaks or final-answer markers. At each step, CaT evaluates the calibrated entropy-based confidence of each active branch, prunes branches whose cumulative confidence falls below $\tau$, and reallocates the remaining budget to more promising branches. 

\begin{algorithm}[t]
\caption{CaT Inference}
\small
\label{alg:cat_inference}
\begin{algorithmic}[1]
   \STATE {\bfseries Input:} Query $x$, calibrated model $\pi_\theta$, threshold $\tau$, branch budget $K$, maximum steps $T$.
   \STATE {\bfseries Initialize:} Active branches $\mathcal{B} \gets \{(\emptyset, 1.0)\}$, completed answers $\mathcal{A} \gets \emptyset$.
   
   \FOR{$t = 1$ {\bfseries to} $T$}
      \STATE $\mathcal{B}_{\mathrm{new}} \gets \emptyset$.
      \FOR{each branch $(z_{1:t-1}, C_{1:t-1}) \in \mathcal{B}$}
          \STATE Generate candidate next steps $\{z_t^{(k)}\}_{k=1}^{K}$ from $\pi_\theta(\cdot \mid x, z_{1:t-1})$.
          \FOR{each candidate step $z_t^{(k)}$}
              \STATE Compute step confidence $c_t^{(k)}$ from calibrated token-level entropy.
              \STATE $C_{1:t}^{(k)} \gets C_{1:t-1} \cdot c_t^{(k)}$.
              \IF{$z_t^{(k)}$ contains a final answer}
                  \STATE Add $(z_{1:t-1}, z_t^{(k)}, C_{1:t}^{(k)})$ to $\mathcal{A}$.
              \ELSIF{$C_{1:t}^{(k)} \ge \tau$}
                  \STATE Add $(z_{1:t-1}, z_t^{(k)}, C_{1:t}^{(k)})$ to $\mathcal{B}_{\mathrm{new}}$.
              \ENDIF
          \ENDFOR
      \ENDFOR
      \STATE Keep the top-$K$ branches in $\mathcal{B}_{\mathrm{new}}$ by cumulative confidence.
      \STATE $\mathcal{B} \gets \mathcal{B}_{\mathrm{new}}$.
      \IF{$\mathcal{B} = \emptyset$}
          \STATE {\bfseries break}
      \ENDIF
   \ENDFOR
   
   \IF{$\mathcal{A} \neq \emptyset$}
      \STATE {\bfseries Return} the completed answer with the highest cumulative confidence.
   \ELSE
      \STATE {\bfseries Return} failure and mark as incorrect under Pass@1.
   \ENDIF
\end{algorithmic}
\end{algorithm}



\end{document}